\documentclass{INTERSPEECH2023}

\interspeechcameraready

\usepackage{cite}
\usepackage{subcaption}
\usepackage{multirow}

\newcommand{\Rb}{\mathbb{R}}

\newcommand{\Av}{\mathbf{A}}
\newcommand{\Bv}{\mathbf{B}}

\newcommand{\Uv}{\mathbf{U}}
\newcommand{\Vv}{\mathbf{V}}

\newcommand{\Xv}{\mathbf{X}}
\newcommand{\Yv}{\mathbf{Y}}
\newcommand{\Zv}{\mathbf{Z}}

\usepackage{xcolor}

\title{A Comparative Study on E-Branchformer vs Conformer\\
in Speech Recognition, Translation, and Understanding Tasks}
\name{Yifan Peng$^1$, Kwangyoun Kim$^2$, Felix Wu$^2$, Brian Yan$^1$, Siddhant Arora$^1$,\\
William Chen$^1$, Jiyang Tang$^1$, Suwon Shon$^2$, Prashant Sridhar$^2$, Shinji Watanabe$^1$}
\address{
  $^1$Carnegie Mellon University, Pittsburgh, PA, USA\\
  $^2$ASAPP Inc., Mountain View, CA, USA}
\email{yifanpen@andrew.cmu.edu, \{kkim,fwu\}@asapp.com}

\begin{document}

\maketitle

\begin{abstract}
Conformer, a convolution-augmented Transformer variant, has become the \textit{de facto} encoder architecture for speech processing due to its superior performance in various tasks, including automatic speech recognition (ASR), speech translation (ST) and spoken language understanding (SLU). Recently, a new encoder called E-Branchformer has outperformed Conformer in the LibriSpeech ASR benchmark, making it promising for more general speech applications. This work compares E-Branchformer and Conformer through extensive experiments using different types of end-to-end sequence-to-sequence models. Results demonstrate that E-Branchformer achieves comparable or better performance than Conformer in almost all evaluation sets across 15 ASR, 2 ST, and 3 SLU benchmarks, while being more stable during training. We will release our training configurations and pre-trained models for reproducibility, which can benefit the speech community.~\footnote{\url{https://github.com/espnet/espnet}}

\end{abstract}

\section{Introduction}

Sequence-to-sequence (seq2seq) models have achieved remarkable success in end-to-end (E2E) speech processing. Various types of neural networks have been explored in recent years, including recurrent neural networks (RNNs)~\cite{listen-attend-spell, zeyer2018improved, asr-s2s}, convolutional neural networks (CNNs)~\cite{cnn-lvcsr, jasper, contextnet} and Transformer networks based on self-attention~\cite{transformer, transformer-vs-rnn, transformer-transducer}. These architectures have complementary capacity in sequence modeling. More recent studies have proposed to combine different networks for better performance. Conformer~\cite{conformer}, a convolution-augmented Transformer encoder, has become the \textit{de facto} standard architecture, due to its superior performance in many speech processing tasks~\cite{conformer-vs-transformer}.
Conformer captures both global and local contexts in a feature sequence through cascaded self-attention and convolution modules. An alternative approach is using parallel branches for different ranged contexts. Branchformer~\cite{branchformer} follows this idea and shows competitive or even better results in several benchmarks for automatic speech recognition (ASR) and spoken language understanding (SLU), while being more stable for training in extreme data regimes. Further, E-Branchformer~\cite{ebranchformer} enhances the vanilla Branchformer with a convolution-based merge module and Conformer-style feed-forward networks. It achieves new state-of-the-art (SOTA) results in the standard LibriSpeech ASR benchmark~\cite{librispeech-corpus}, which is highly encouraging for general speech applications.

In this work, we explore the efficacy of E-Branchformer in various speech processing tasks, including ASR, speech translation (ST) and SLU. We compare E-Branchformer versus Conformer through extensive experiments in a wide range of publicly available benchmarks (i.e., 15 ASR, 2 ST, and 3 SLU corpora). We also investigate different E2E frameworks, including connectionist temporal classification (CTC)~\cite{ctc}, attention-based encoder-decoder (AED) and RNN-transducer (RNN-T)~\cite{rnnt}. Results show that E-Branchformer performs equally well as or better than Conformer in almost all evaluation sets. E-Branchformer is also more stable to train when the model is large or the dataset is small. We share various training tips based on our investigations. We will also release our training configurations and pre-trained models for full reproducibility, which can significantly benefit the speech community.

\section{Models}

We follow the default setup in the open-source ESPnet toolkit~\cite{espnet, espnet-st, espnet-slu}. Only the encoder is changed to compare E-Branchformer~\cite{ebranchformer} and Conformer~\cite{conformer}, while the other components are the same. In a speech encoder, the raw audio waveform is first processed by a frontend to extract speech features such as log Mel filterbanks or self-supervised learning (SSL) based features. The feature sequence is then fed into a convolutional subsampling module. The downsampled feature sequence is further processed by a stack of identical encoder layers to capture high-level contextual information. Depending on the E2E framework, the final output sequence can be fed into a Transformer decoder, an RNN-T joint network or a CTC layer.

\subsection{Conformer}

\begin{figure}[t]
     \begin{subfigure}[b]{0.45\linewidth}
          \centering
          \includegraphics[width=0.72\linewidth]{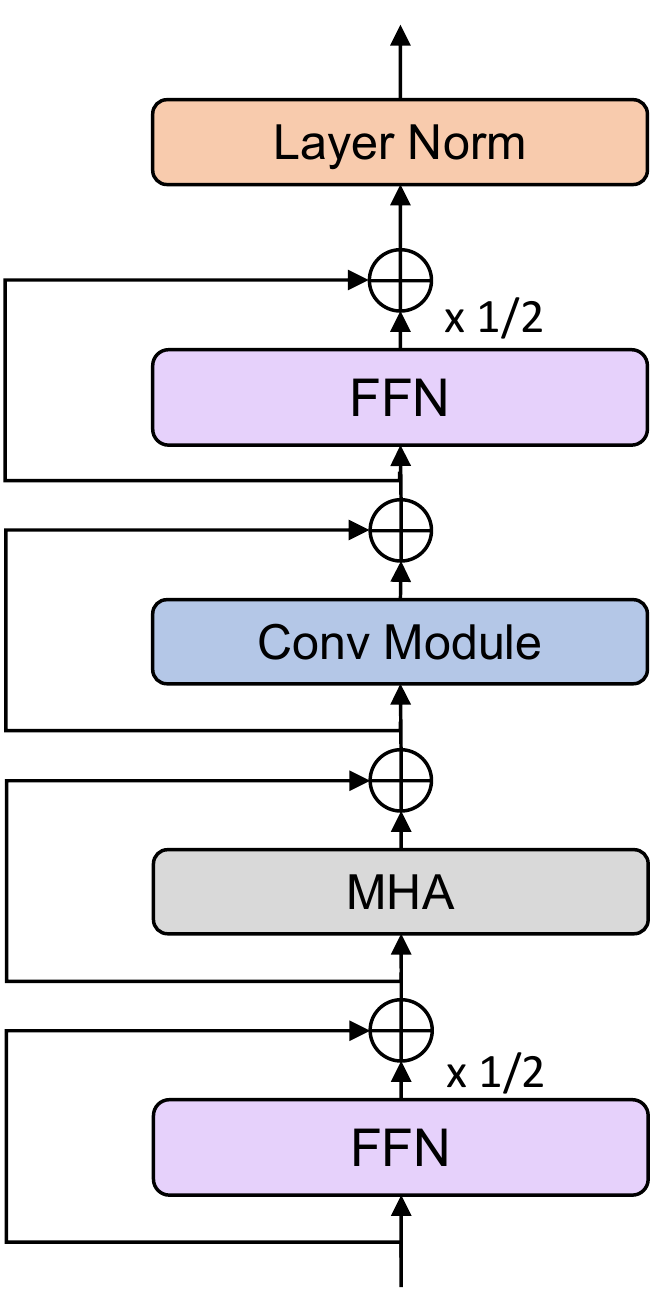}
          \caption{Conformer~\cite{conformer}}
          \label{fig:conformer}
     \end{subfigure}
     \hfill
     \begin{subfigure}[b]{0.45\linewidth}
          \centering
          \includegraphics[width=0.82\linewidth]{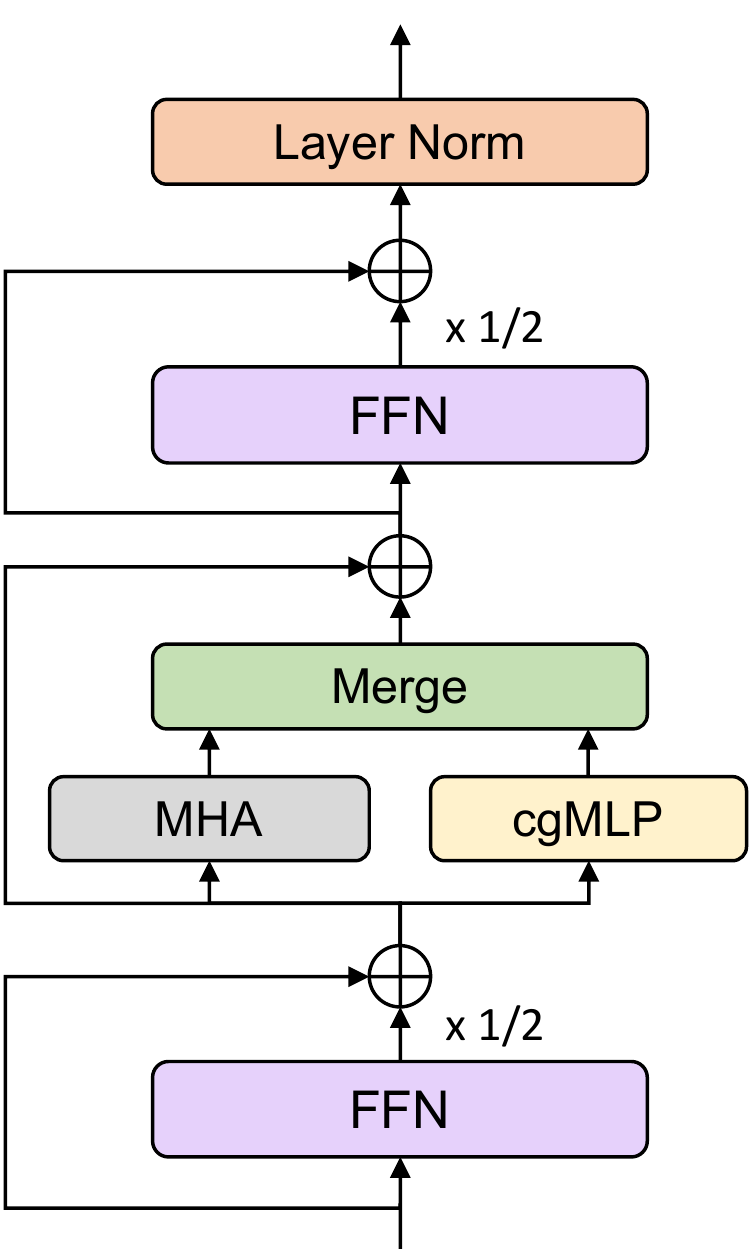}
          \caption{E-Branchformer~\cite{ebranchformer}}
          \label{fig:ebranch}
     \end{subfigure}
     \vskip -0.1in
     \caption{Comparison of encoder architectures.}
     \label{fig:arch}
    \vskip -0.2in
\end{figure}

Figure~\ref{fig:conformer} illustrates a Conformer layer~\cite{conformer}, which consists of a position-wise feed-forward network (FFN), a multi-head self-attention (MHA) module, a convolution (Conv) module and another FFN. Each of these modules has a residual connection~\cite{resnet} and a layer normalization~\cite{layernorm} in a pre-norm style~\cite{wang-etal-2019-learning-deep, nguyen-salazar-2019-transformers}. Different modules are combined sequentially. 
For an input sequence $\Xv \in\Rb^{T\times d}$ of length $T$ and feature size $d$, the final output of a Conformer layer $\Yv_{\text{conf}} \in \Rb^{T\times d}$ is computed as follows:
\begin{align}
    \Xv^1_{\text{conf}} & = \Xv + \frac{1}{2} \text{FFN}^1(\Xv), \\
    \Xv^2_{\text{conf}} & = \Xv^1_{\text{conf}} + \text{MHA}(\Xv^1_{\text{conf}}), \\
    \Xv^3_{\text{conf}} & = \Xv^2_{\text{conf}} + \text{Conv}(\Xv^2_{\text{conf}}), \\
    \Xv^4_{\text{conf}} & = \Xv^3_{\text{conf}} + \frac{1}{2} \text{FFN}^2(\Xv^3_{\text{conf}}), \\
    \Yv_{\text{conf}} & = \text{LayerNorm}(\Xv^4_{\text{conf}}),
\end{align}
where $\Xv^1_{\text{conf}}, \Xv^2_{\text{conf}}, \Xv^3_{\text{conf}}, \Xv^4_{\text{conf}} \in \Rb^{T\times d}$ are intermediate outputs. Each FFN is composed of two linear projections and a Swish activation~\cite{swish} in between. Two separate FFNs with half-step residual weights are employed as in Macaron-Net~\cite{macaron-net}. Unlike the vanilla Transformer~\cite{transformer}, MHA uses relative positional encodings from Transformer-XL~\cite{transformer-xl}. The key component of Conformer is the Conv module which contains a point-wise convolution followed by a gated linear unit (GLU) activation~\cite{glu}, a 1-D depth-wise convolution, a batch normalization~\cite{batchnorm}, a Swish activation and another point-wise convolution.

\subsection{E-Branchformer}

\begin{figure}[t]
    \centering
    \includegraphics[width=0.43\linewidth]{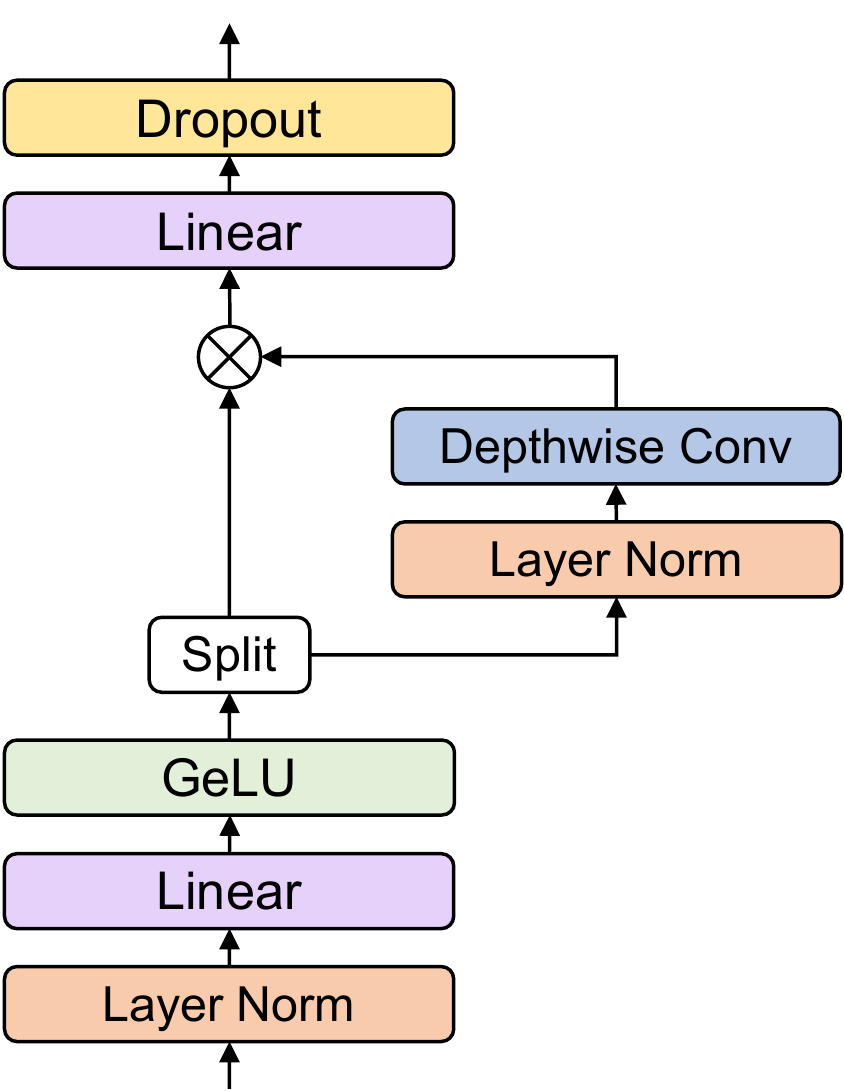}
    \vskip -0.1in
    \caption{Architecture of cgMLP~\cite{cgmlp}.}
    \label{fig:cgmlp}
    \vskip -0.2in
\end{figure}

E-Branchformer~\cite{ebranchformer} is an enhanced version of Branchformer~\cite{branchformer}. Figure~\ref{fig:ebranch} shows its architecture. Similar to Conformer, it also has two Macaron-style FFNs. The difference is that E-Branchformer contains two parallel branches between the FFNs as proposed in Branchformer~\cite{branchformer}. One branch captures global context using MHA while the other branch captures local context using multi-layer perceptron with convolutional gating (cgMLP)~\cite{cgmlp}. Two branches are merged by a concatenation operation, a 1-D depth-wise convolution and a linear projection, which is more effective than the simple concatenation followed by a linear projection used in Branchformer. For input $\Xv \in \Rb^{T\times d}$, the output $\Yv_{\text{ebf}} \in \Rb^{T \times d}$ is defined as follows:
\begin{align}
    \Xv^1_\text{ebf} & = \Xv + \frac{1}{2} \text{FFN}^1(\Xv),\\
    \Xv^{2,\text{mha}}_\text{ebf}, ~~ \Xv^{2, \text{mlp}}_\text{ebf} &= \text{MHA}(\Xv^1_\text{ebf}), ~~ \text{cgMLP}(\Xv^1_\text{ebf}), \\
    \Xv^2_\text{ebf} & = \Xv^1_\text{ebf} + \text{Merge}(\Xv^{2,\text{mha}}_\text{ebf}, \Xv^{2, \text{mlp}}_\text{ebf}), \\
    \Xv^3_\text{ebf} & = \Xv^2_\text{ebf} + \frac{1}{2}\text{FFN}^2(\Xv^2_\text{ebf}), \\
    \Yv_\text{ebf} & = \text{LayerNorm}(\Xv^3_\text{ebf}),
\end{align}
where $\Xv^1_\text{ebf}, \Xv^{2,\text{mha}}_\text{ebf}, \Xv^{2, \text{mlp}}_\text{ebf}, \Xv^2_\text{ebf}, \Xv^3_\text{ebf} \in \Rb^{T\times d}$ are intermediate outputs.
Figure~\ref{fig:cgmlp} shows the architecture of cgMLP~\cite{cgmlp}, which leverages depth-wise convolution and linear gating to extract local contextual information. For input $\Xv^1_\text{ebf} \in \Rb^{T\times d}$, the output $\Xv^{2, \text{mlp}}_\text{ebf}$ is derived as follows:
\begin{align}
    \Zv &= \text{GeLU}(\text{LayerNorm}(\Xv^1_\text{ebf}) \Uv) \in \Rb^{T \times d_{\text{mlp}}}, \label{eq:cgmlp1} \\
    \Av, \Bv &= \text{Split}(\Zv) \in \Rb^{T \times \frac{1}{2} d_{\text{mlp}}}, \\
    \tilde{\Zv} &= \Av \odot \text{DwConv}(\text{LayerNorm}(\Bv)) \in \Rb^{T \times \frac{1}{2} d_{\text{mlp}}}, \\
    \Xv^{2, \text{mlp}}_\text{ebf} &= \text{Dropout}(\tilde{\Zv} \Vv) \in \Rb^{T\times d},
\end{align}
where $\Uv \in \Rb^{d \times d_\text{mlp}}$ and $\Vv \in \Rb^{\frac{1}{2} d_\text{mlp} \times d}$ are two linear projections. $\odot$ denotes element-wise product.

\section{Speech recognition experiments}

\subsection{Setups}

\noindent\textbf{Data.}
A total of 15 public ASR datasets are utilized, covering various languages (English, Chinese, Spanish, Japanese, Italian, or even multilingual with 102 languages), recording environments (clean, noisy, far-field), speech types (spontaneous, read), and sizes (10 to 10k hours). Disordered speech from AphasiaBank~\cite{aphasiabank} is also evaluated. The evaluation metric is character error rate (CER) or word error rate (WER). The total model size and encoder's multiply-accumulate operations (MACs) for a 10-second audio are also reported.

\noindent\textbf{Models.}
We mainly use the attention-based encoder-decoder (AED) with joint CTC training and decoding~\cite{joint-ctc-att,joint-ctc-att-decoding}. The encoder is either Conformer or E-Branchformer, while the decoder is a 6-layer Transformer. Log Mel filterbanks are extracted by default, except that FLEURS uses an SSL frontend as in~\cite{william-fleurs}. FLEURS also exploits intermediate CTC~\cite{interctc} and self-condition CTC~\cite{self-condition-ctc} in the encoder. We also conduct experiments using pure CTC or RNN-T models in a subset of datasets.

\noindent\textbf{Training.}
We follow the ESPnet2 recipes\footnote{\url{https://github.com/espnet/espnet/tree/master/egs2}} for data preparation, model training and decoding. Most recipes perform speed perturbation with ratios $\{0.9, 1.0, 1.1\}$ and SpecAugment~\cite{specaugment}. A medium-sized model with hidden size $d=256$ is employed by default. For FLEURS, GigaSpeech and LibriSpeech 960h, a larger model with $d=512$ is trained instead. The Adam optimizer~\cite{adam} with warmup learning rate schedule~\cite{transformer} is employed. Training hyperparameters such as the learning rate, weight decay and warmup steps are from existing baselines. We will release our detailed setups to ensure reproducibility.

\subsection{Results}

\begingroup
\setlength{\tabcolsep}{2pt}

\begin{table*}[t]
  \caption{CER or WER (\%) on speech recognition benchmarks using attention-based encoder-decoder (AED) with joint CTC. The total number of parameters ($\times 10^6)$ and encoder's multiply-accumulate (MAC) operations ($\times 10^9$) are also reported. $\dagger$ means a frozen SSL frontend is used but not counted. $\ddagger$ means a language model is used with shallow fusion following existing ESPnet2 recipes.}
  \label{tab:main-results}
  \vskip -0.1in
  \centering
  \resizebox {0.97\linewidth} {!} {
  \begin{tabular}{c @{\hskip 2pt} c @{\hskip 2pt} c|c|c|c|c|c|c|c}
  \toprule
  \multirow{2}{*}{Dataset} & \multirow{2}{*}{Token} & \multicolumn{1}{c}{\multirow{2}{*}{Metric}} & \multicolumn{1}{c}{\multirow{2}{*}{Evaluation Sets}} & \multicolumn{3}{c}{Conformer} & \multicolumn{3}{c}{E-Branchformer}\\
  \cmidrule(lr){5-7}
  \cmidrule(lr){8-10}
  & & \multicolumn{1}{c}{} & \multicolumn{1}{c}{} & \multicolumn{1}{c}{Params} & \multicolumn{1}{c}{MACs} & \multicolumn{1}{c}{Results $\downarrow$} & \multicolumn{1}{c}{Params} & \multicolumn{1}{c}{MACs} & \multicolumn{1}{c}{Results $\downarrow$} \\
  \midrule
  \midrule
  AIDATATANG~\cite{aidatatang} & Char & CER & dev / test & 46.0 & 14.7 & $\ddagger$ 3.6 / 4.3 & 45.4 & 15.5 & $\ddagger$ \textbf{3.4} / \textbf{4.1} \\
  AISHELL~\cite{aishell-corpus} & Char & CER & dev / test & 46.3 & 15.3 & 4.3 / 4.6 & 45.7 & 15.5 & \textbf{4.2} / \textbf{4.4} \\
  AphasiaBank~\cite{aphasiabank} & Char & WER & patients / control & 44.2 & 30.1 & 40.3 / 35.3 & 45.7 & 32.0 & \textbf{36.2} / \textbf{31.2} \\
  CHiME4~\cite{chime4} & Char & WER & \{dt05,et05\}\_\{simu,real\} & 30.4 & 8.8 & $\ddagger$ 7.8 / 9.5 / 12.5 / 14.8 & 30.8 & 8.8 & $\ddagger$ \textbf{6.8} / \textbf{8.4} / \textbf{10.8} / \textbf{13.0} \\
  Fisher-Callhome~\cite{fisher-callhome} & BPE & WER & dev / dev2 / test / devtest / evltest & 43.8 & 11.6 & 20.7 / 20.9 / 19.4 / 38.3 / 38.8 & 43.2 & 12.1 & \textbf{20.5} / \textbf{20.2} / \textbf{18.7} / \textbf{37.8} / \textbf{37.6} \\
  FLEURS~\cite{FLEURS} & BPE & CER & dev / test & $\dagger$126.6 & 48.8 & $\ddagger$ 10.1 / 10.4 & $\dagger$127.4 & 50.1 & $\ddagger$ \textbf{9.3} / \textbf{9.2} \\
  GigaSpeech~\cite{gigaspeech} & BPE & WER & dev / test & 116.2 & 20.0 & 10.9 / 10.8 & 148.9 & 26.1 & \textbf{10.6} / \textbf{10.5} \\
  JSUT~\cite{jsut} & Char & CER & dev / eval1 & 45.1 & 11.6 & $\ddagger$ 12.3 / 13.6 & 44.2 & 12.1 & $\ddagger$ \textbf{11.8} / \textbf{13.0} \\
  LibriSpeech 100h~\cite{librispeech-corpus} & BPE & WER & \{dev,test\}\_\{clean,other\} & 39.0 & 10.3 & 6.3 / 17.0 / 6.6 / 17.2 & 38.5 & 9.9 & \textbf{6.1} / \textbf{16.7} / \textbf{6.3} / \textbf{17.0} \\
  LibriSpeech 960h~\cite{librispeech-corpus} & BPE & WER & \{dev,test\}\_\{clean,other\} & 147.8 & 42.5 & $\ddagger$ 1.72 / 3.65 / \textbf{1.85} / 3.95 & 148.9  & 42.7 & $\ddagger$ \textbf{1.67} / \textbf{3.64} / \textbf{1.85} / \textbf{3.71} \\
  MuST-C~\cite{must-c} & BPE & WER & tst-\{COMMON, HE\}.en-de & 46.1 & 12.0 & 7.7 / 6.7 & 37.7 & 9.9 & \textbf{7.3} / \textbf{6.0} \\
  Switchboard~\cite{swbd-corpus} & BPE & WER & eval2000 (callhm / swbd) & 36.7 & 10.3 & 13.5 / 7.4 & 36.2 & 9.9 & \textbf{13.4} / \textbf{7.3} \\
  TEDLIUM2~\cite{tedlium2} & BPE & WER & dev / test & 35.5 & 10.3 & 7.5 / 7.6 & 35.0 & 9.9 & \textbf{7.3} / \textbf{7.1} \\
  VoxForge~\cite{voxforge} & Char & CER & dt\_it / et\_it & 35.2 & 13.2 & 9.0 / 8.1 & 34.7 & 12.6 & \textbf{8.8} / \textbf{8.0} \\
  WSJ~\cite{wsj} & Char & WER & dev93 / eval92 & 35.2 & 13.2 & $\ddagger$ \textbf{6.5} / \textbf{4.1} & 34.7 & 12.6 & $\ddagger$ \textbf{6.5} / 4.3 \\
  \bottomrule
  \end{tabular}
  }
  \vskip -0.1in
\end{table*}
\endgroup

\begingroup
\setlength{\tabcolsep}{2pt}

\begin{table}[t]
  \caption{CER or WER (\%) of different configurations using AED with joint CTC. ``Conformer-Deep'' has 15 encoder layers with 1024 FFN units, while ``Conformer-Wide'' has 12 encoder layers with 2048 FFN units. Evaluation sets are the same as in Table~\ref{tab:main-results}, except that LibriSpeech 100h only shows test sets.}
  \label{tab:diff-configs}
  \vskip -0.1in
  \centering
  \resizebox {\linewidth} {!} {
  \begin{tabular}{c|c|c|c|c|c|c}
    \toprule
    \multicolumn{1}{c}{\multirow{2}{*}{Dataset}} & \multicolumn{2}{c}{Conformer-Deep} & \multicolumn{2}{c}{Conformer-Wide} & \multicolumn{2}{c}{E-Branchformer}\\
    \cmidrule(lr){2-3}
    \cmidrule(lr){4-5}
    \cmidrule(lr){6-7}
    \multicolumn{1}{c}{} & \multicolumn{1}{c}{Params} & \multicolumn{1}{c}{Results $\downarrow$} & \multicolumn{1}{c}{Params} & \multicolumn{1}{c}{Results $\downarrow$} & \multicolumn{1}{c}{Params} & \multicolumn{1}{c}{Results $\downarrow$} \\
    \midrule
    \midrule
    LibriSpeech 100h & 39.0 & 6.6 / 17.2 & 46.8 & 6.6 / 17.1 & 38.5 & \textbf{6.3} / \textbf{17.0} \\
    Switchboard & 36.7 & 13.5 / 7.4 & 44.5 & 13.8 / 7.5 & 36.2 & \textbf{13.4} / \textbf{7.3} \\
    TEDLIUM2 & 35.5 & 7.5 / 7.6 & 43.4 & 7.5 / 7.5 & 35.0 & \textbf{7.3} / \textbf{7.1} \\
    VoxForge & 35.2 & 9.0 / 8.1 & 43.0 & 8.9 / \textbf{8.0} & 34.7 & \textbf{8.8} / \textbf{8.0} \\
    \bottomrule
  \end{tabular}
  }
    \vskip -0.05in
\end{table}
\endgroup

\begingroup
\setlength{\tabcolsep}{2pt}

\begin{table}[t]
  \caption{CER or WER (\%) of pure CTC-based models. Greedy search is performed without a language model.}
  \label{tab:ctc}
  \vskip -0.1in
  \centering
  \resizebox {\linewidth} {!} {
  \begin{tabular}{c|c|c|c|c}
    \toprule
    \multicolumn{1}{c}{\multirow{2}{*}{Dataset}} & \multicolumn{2}{c}{Conformer} & \multicolumn{2}{c}{E-Branchformer}\\
    \cmidrule(lr){2-3}
    \cmidrule(lr){4-5}
    \multicolumn{1}{c}{} & \multicolumn{1}{c}{Params} & \multicolumn{1}{c}{Results $\downarrow$} & \multicolumn{1}{c}{Params} & \multicolumn{1}{c}{Results $\downarrow$} \\
    \midrule
    \midrule
    AISHELL & 26.8 & 5.8 / 6.3 & 26.2 & \textbf{5.4} / \textbf{6.0} \\
    LibriSpeech 100h & 27.0 & 9.4 / 22.5 / 9.9 / \textbf{23.1} & 26.4 & \textbf{9.2} / \textbf{22.4} / \textbf{9.6} / \textbf{23.1} \\
    TEDLIUM2 & 25.8 & 9.1 / 9.0 & 25.3 & \textbf{8.7} / \textbf{8.3} \\
    \bottomrule
  \end{tabular}
  }
  \vskip -0.12in
\end{table}
\endgroup

\begingroup
\setlength{\tabcolsep}{2pt}

\begin{table}[t]
  \caption{CER or WER (\%) of RNN-T models. The decoder is a single-layer LSTM~\cite{lstm}. Beam search with beam size 10 is performed without a language model.}
  \label{tab:transducer}
  \vskip -0.05in
  \centering
  \resizebox {\linewidth} {!} {
  \begin{tabular}{c|c|c|c|c}
    \toprule
    \multicolumn{1}{c}{\multirow{2}{*}{Dataset}} & \multicolumn{2}{c}{Conformer} & \multicolumn{2}{c}{E-Branchformer}\\
    \cmidrule(lr){2-3}
    \cmidrule(lr){4-5}
    \multicolumn{1}{c}{} & \multicolumn{1}{c}{Params} & \multicolumn{1}{c}{Results $\downarrow$} & \multicolumn{1}{c}{Params} & \multicolumn{1}{c}{Results $\downarrow$} \\
    \midrule
    \midrule
    AISHELL & 29.9 & \textbf{4.9} / 5.3 & 29.4 & \textbf{4.9} / \textbf{5.2} \\
    LibriSpeech 100h & 30.5 & \textbf{6.6} / 17.9 / 6.9 / 18.1 & 30.0 & \textbf{6.6} / \textbf{17.6} / \textbf{6.8} / \textbf{18.0} \\
    TEDLIUM2 & 26.8 & 8.1 / 7.7 & 26.3 & \textbf{7.6} / \textbf{7.4} \\
    \bottomrule
  \end{tabular}
  }
    \vskip -0.1in
\end{table}
\endgroup

Table~\ref{tab:main-results} summarizes the ASR results of AED models with joint CTC, which is the most widely used setup in ESPnet. Compared to those well-established Conformer baselines, E-Branchformer achieves comparable or superior results with a similar model size and computational complexity in almost all benchmarks. \textbf{We have only observed a slight degradation in one set among 39 evaluation sets across 15 corpora.} The improvements are especially remarkable in AphasiaBank, CHiME4, Fisher-Callhome, FLEURS, JSUT, MuST-C and TEDLIUM2, indicating that E-Branchformer has strong modeling capacities for various speech types.

Conformer configurations can vary across different datasets, with some datasets benefiting from deeper networks while others may benefit from wider networks.
Table~\ref{tab:diff-configs} compares E-Branchformer with two Conformer baselines in four benchmarks. The hidden sizes are $d=256$ for all models. E-Branchformer consists of 12 encoder layers with 1024 FFN units and 1024 MLP units ($d_\text{mlp}$ in Eq.~\eqref{eq:cgmlp1}). ``Conformer-Deep'' has 15 encoder layers with 1024 FFN units, while ``Conformer-Wide'' has 12 layers with 2048 FFN units. E-Branchformer consistently achieves lower CERs or WERs than ``Conformer-Deep'' with a similar size. It even shows better or similar performance than ``Conformer-Wide'' whose size is 20\% larger.

Different E2E ASR frameworks (i.e., AED, CTC and RNN-T) typically share a similar encoder. To investigate the generalizability of Conformer and E-Branchformer encoders, we apply them to pure CTC and RNN-T models. Table~\ref{tab:ctc} and Table~\ref{tab:transducer} summarize the two sets of experiments, respectively. Similar to the AED results, E-Branchformer achieves consistent improvements over Conformer in various evaluation sets under the same training condition. This demonstrates that E-Branchformer encoders are capable of extracting better contextualized speech representations which generally benefit E2E ASR.

\subsection{Discussions}

The following are some observations and training tips based on our investigations.

\begin{itemize}
    \item When the model is large or the data is small, E-Branchformer (and Branchformer) can offer greater training stability than Conformer, as observed empirically. For instance, in an experiment with TEDLIUM2 where we set the peak learning rate to 2e-3 and used pure CTC with 33M parameter encoders, \textbf{Conformer failed in 6 out of 10 random trials, while E-Branchformer only failed twice.} One reason for this may be that Conformer's sequential combination of modules increases the encoder's depth and makes it harder to converge. Moreover, in the successful trials, E-Branchformer exhibited lower validation loss in an early stage of training. It is worth noting that although a lower learning rate could improve training stability, in our case, it degraded the WERs.
    \item A similar configuration of E-Branchformer performs well in many ASR corpora. Using Macaron-style FFNs like those in Conformer is generally beneficial. For medium-sized models, we recommend setting the hidden size to $d=256$ and using $4\times$ expansion in FFNs and cgMLP. For large models, we suggest following the original paper~\cite{ebranchformer} and using $d=512$ with $6\times$ expansion in cgMLP and $2\times$ expansion in FFNs.
    \item The same training hyperparameters (e.g., batch size, learning rate, warmup steps, total epochs) used by Conformer usually work well for E-Branchformer assuming their sizes are close. In most experiments, we did not change these configurations.
    \item Stochastic depth~\cite{stochastic-depth} is disabled in most of our experiments for fair comparison with prior baselines. However, we do find that it can slightly improve the final results for datasets like FLEURS and LibriSpeech 960h. Moreover, it can be applied to both encoder and decoder, e.g., with dropout probabilities 0.1 and 0.2, respectively.
    \item With joint CTC training~\cite{joint-ctc-att}, the auxiliary CTC loss can be a reliable metric for assessing the success of the training process in an early stage. If the validation loss does not decrease in the first few epochs, we need to reduce the learning rate or increase warmup steps. This also applies to Conformer.
\end{itemize}

\section{Speech translation experiments}

\subsection{Setups}

\noindent\textbf{Data.}
Two ST datasets are used. MuST-C \cite{must-c} is a TED-Talk corpus for English-to-X translation. We use the 400 hour v2 set of English-to-German.
Fisher-Callhome \cite{fisher-callhome} is a 170 hour conversational corpus for Spanish-to-English translation.

\noindent\textbf{Models.}
We use the attention-based encoder-decoder (AED) with joint CTC training and decoding~\cite{joint-ctc-att,joint-ctc-att-decoding}. 
Our ST models are similar to our ASR models, except they use hierarchical CTC encoders \cite{yan2022ctc} which consist of a 12-layer Conformer or E-Branchformer attached to an ASR CTC criterion followed by another 6-layer Conformer or 8-layer E-Branchformer attached to an ST CTC criterion.
Given the greater number of layers for ST models, we use a slightly higher number of layers for our E-Branchformer models to keep parameter counts in a similar range -- E-Branchformer models are still smaller.
ST models use ASR pre-training for the first 12 encoder layers.

\noindent\textbf{Training.}
We follow the ESPnet2 recipes for data preparation, model training and decoding. Speed perturbation with ratios $\{0.9, 1.0, 1.1\}$ and SpecAugment~\cite{specaugment} are performed. The medium-sized model with hidden size $d=256$ is used. The Adam~\cite{adam} optimizer with warmup~\cite{transformer} is employed.

\subsection{Results}

Table~\ref{tab:st} shows the ST results. E-Branchformer achieves a higher BLEU score than Conformer on Callhome (21.9 vs. 21.2) and also shows minor improvements on MuST-C and Fisher with a slightly smaller model size.
This suggests that E-Branchformer is capable of handling non-monotonic sequence transductions such as translation where source-to-target word re-ordering may occur.

\begingroup
\setlength{\tabcolsep}{5pt}

\begin{table}[t]
  \caption{Speech translation results.}
  \label{tab:st}
  \vskip -0.1in
  \centering
  \resizebox {0.8\linewidth} {!} {
  \begin{tabular}{c|c|c|c|c}
    \toprule
    \multicolumn{1}{c}{\multirow{2}{*}{Dataset}} & \multicolumn{2}{c}{Conformer} & \multicolumn{2}{c}{E-Branchformer}\\
    \cmidrule(lr){2-3}
    \cmidrule(lr){4-5}
    \multicolumn{1}{c}{} & \multicolumn{1}{c}{Params} & \multicolumn{1}{c}{BLEU $\uparrow$} & \multicolumn{1}{c}{Params} & \multicolumn{1}{c}{BLEU $\uparrow$} \\
    \midrule
    \midrule
    MuST-C~\cite{must-c} & 74.5 & 28.6  & 71.4 & \textbf{28.7} \\
    Fisher~\cite{fisher-callhome} & 69.8 & 55.5 & 66.8 & \textbf{55.6} \\
    Callhome~\cite{fisher-callhome} & 69.8 & 21.2 & 66.8 & \textbf{21.9} \\
    \bottomrule
  \end{tabular}
  }
  \vskip -0.17in
\end{table}
\endgroup

\section{Speech understanding experiments}

\subsection{Setups}

\noindent\textbf{Data.}
Three SLU datasets are used. SLURP~\cite{slurp-corpus} is a multi-domain corpus for intent classification and entity recognition, which contains single-turn user interactions with a home assistant. SLUE~\cite{shon2022slue} is a low-resource benchmark containing naturally produced speech for named entity recognition (NER) and sentiment analysis. STOP~\cite{stop} is a large-scale corpus for spoken task-oriented semantic parsing.

\noindent\textbf{Models.}
As in ESPnet-SLU~\cite{espnet-slu}, SLU tasks are formulated as seq2seq problems. The input is a sequence of speech features, and the output is a sequence of text tokens including special SLU labels. Then, the same E2E ASR models can be applied. Specifically, we employ AED models with joint CTC~\cite{joint-ctc-att, joint-ctc-att-decoding}, where the decoder is a 6-layer Transformer. Similar to~\cite{pengIntegrationSLU}, an SSL frontend is used for SLUE and STOP.

\noindent\textbf{Training.}
We follow the ESPnet2 recipes for data preparation, training and decoding. Speed perturbation and SpecAugment~\cite{specaugment} are performed for data augmentation. The Adam~\cite{adam} optimizer with warmup~\cite{transformer} is employed.

\subsection{Results}

Table~\ref{tab:slu} shows SLU results. E-Branchformer has superior performance over Conformer on SLURP with a similar model size. It also achieves a higher accuracy on STOP, but the model size is larger due to the reuse of configuration from LibriSpeech 960h. For low-resource SLUE-voxpopuli (NER) and SLUE-voxceleb (sentiment analysis) corpora, training is done 3 times with different random seeds, and metrics are averaged. We observe that E-Branchformer is slightly better on SLUE-voxpopuli but worse on SLUE-voxceleb. This indicates that the frozen SSL frontend might be more important than the additional encoder layers in low-resource SLU tasks.

\begingroup
\setlength{\tabcolsep}{4pt}

\begin{table}[t]
  \caption{Spoken language understanding results on test sets. SLURP shows intent classification accuracy (\%) and SLU-F1 (\%). SLUE-voxpopuli shows micro and macro F1 (\%). SLUE-voxceleb shows macro F1 (\%). STOP shows exact match accuracy (\%). $\dagger$ means a frozen SSL frontend is used but not counted.}
  \label{tab:slu}
  \vskip -0.1in
  \centering
  \resizebox {0.9\linewidth} {!} {
  \begin{tabular}{c|c|c|c|c}
    \toprule
    \multicolumn{1}{c}{\multirow{2}{*}{Dataset}} & \multicolumn{2}{c}{Conformer} & \multicolumn{2}{c}{E-Branchformer}\\
    \cmidrule(lr){2-3}
    \cmidrule(lr){4-5}
    \multicolumn{1}{c}{} & \multicolumn{1}{c}{Params} & \multicolumn{1}{c}{Results $\uparrow$} & \multicolumn{1}{c}{Params} & \multicolumn{1}{c}{Results $\uparrow$} \\
    \midrule
    \midrule
    SLURP~\cite{slurp-corpus} & 109.4 & 86.5 / 76.9 & 110.2 & \textbf{87.4} / \textbf{77.6} \\
    SLUE-voxpopuli~\cite{shon2022slue} & $\dagger$ 32.4 & 68.6 / 55.8 & $\dagger$ 33.5 & \textbf{68.7 / 55.9}\\
    SLUE-voxceleb~\cite{shon2022slue} & $\dagger$ 32.4 & \textbf{38.5} & $\dagger$ 33.5 & 38.1\\
    STOP~\cite{stop} & $\dagger$ 114.3 & 73.2 & $\dagger$ 146.8 & \textbf{74.0} \\
    \bottomrule
  \end{tabular}
  }
  \vskip -0.2in
\end{table}
\endgroup

\section{Conclusion}
This work investigates the effectiveness of E-Branchformer in various speech processing tasks, including ASR, ST, and SLU, and compares it to Conformer using different E2E frameworks.
Our extensive experiments in publicly available benchmarks have shown that E-Branchformer outperforms Conformer in a wide variety of tasks, and can be more stable when training with large model size or on small datasets. In addition, we share various training tips and we will release our training configurations and pre-trained models to ensure full reproducibility, which can greatly benefit the speech community.
Future research directions would include evaluating E-Branchformer on more diverse and challenging datasets in low-resource languages and noisy environments. Additionally, the E-Branchformer architecture may also improve the performance of SSL models such as WavLM~\cite{wavlm} and data2vec~2.0~\cite{data2vec2}.

\section{Acknowledgements}
This work used PSC Bridges2 and NCSA Delta through allocation CIS210014 from the Advanced Cyberinfrastructure Coordination Ecosystem: Services \& Support (ACCESS) program, which is supported by National Science Foundation grants \#2138259, \#2138286, \#2138307, \#2137603, and \#2138296.

\bibliographystyle{IEEEtran}
\bibliography{mybib}

\end{document}